\documentclass[final]{cvpr}

\usepackage{times}
\usepackage{epsfig}
\usepackage{graphicx}
\usepackage{amsmath}
\usepackage{amssymb}
\usepackage[utf8]{inputenc} 
\usepackage[T1]{fontenc}    
\usepackage{url}            
\usepackage{booktabs}       
\usepackage{amsfonts}       
\usepackage{amsmath}
\usepackage{nicefrac}       
\usepackage{microtype}      
\usepackage[ruled,vlined]{algorithm2e}
\usepackage{subfigure}
\usepackage{multirow}
\usepackage{aecompl}

\usepackage[pagebackref=true,breaklinks=true,colorlinks,bookmarks=false]{hyperref}

\def\cvprPaperID{6624} 
\def\confYear{CVPR 2021}

\begin{document}

\title{PWCLO-Net: Deep LiDAR Odometry in 3D Point Clouds Using Hierarchical Embedding Mask Optimization}


	\author{Guangming~Wang\textsuperscript{\rm 1}\qquad~Xinrui~Wu\textsuperscript{\rm 1}\qquad~Zhe~Liu\textsuperscript{\rm 2}\qquad~Hesheng~Wang\textsuperscript{\rm 1}\thanks{ Corresponding Author. The first three authors contributed equally.
	}\\
	{\textsuperscript{\rm 1}Department of Automation, Insititue of Medical Robotics, Key Laboratory of System Control}
		\\
		{and Information Processing of Ministry of Education, Shanghai Jiao Tong University}

		  \\
	{\textsuperscript{\rm 2}Department of Computer Science and Technology,
		University of Cambridge} ~~
 \\
	\small{\texttt{\{wangguangming,916806487,wanghesheng\}@sjtu.edu.cn}} \qquad
	\small{\texttt{zl457@cam.ac.uk}}
}

\maketitle
\thispagestyle{empty}
\pagestyle{empty}

\vspace{-0.5cm}
\begin{abstract}
\vspace{-0.2cm}
      A novel 3D point cloud learning model for deep LiDAR odometry, named PWCLO-Net, using hierarchical embedding mask optimization is proposed in this paper. In this model, the Pyramid, Warping, and Cost volume (PWC) structure for the LiDAR odometry task is built to refine the estimated pose in a coarse-to-fine approach hierarchically. An attentive cost volume is built to associate two point clouds and obtain embedding motion patterns. Then, a novel trainable embedding mask is proposed to weigh the local motion patterns of all points to regress the overall pose and filter outlier points. The estimated current pose is used to warp the first point cloud to bridge the distance to the second point cloud, and then the cost volume of the residual motion is built. At the same time, the embedding mask is optimized hierarchically from coarse to fine to obtain more accurate filtering information for pose refinement. The trainable pose warp-refinement process is iteratively used to make the pose estimation more robust for outliers. The superior performance and effectiveness of our LiDAR odometry model are demonstrated on KITTI odometry dataset. Our method outperforms all recent learning-based methods and outperforms the geometry-based approach, LOAM with mapping optimization, on most sequences of KITTI odometry dataset.Our source codes will be released on https://github.com/IRMVLab/PWCLONet.
\end{abstract}
\vspace{-0.7cm}
\section{Introduction}
\vspace{-4pt}
The visual/LiDAR odometry is one of the key technologies in autonomous driving. This task uses two consecutive images or point clouds to obtain the relative pose transformation between two frames, and acts as the base of the subsequential planning and decision making of mobile robots \cite{robot_}. Recently, learning-based odometry methods have shown impressive accuracy on datasets compared with conventional methods based on hand-crafted features. It is found that learning-based methods can deal with sparse features and dynamic environments \cite{dasgil,lpdnet}, which are usually difficult for conventional methods. To our knowledge, most learning-based methods are on the 2D visual odometry \cite{wang2017deepvo,zhou2017unsupervised,min2020voldor,yang2020d3vo,Unsupervised_depth,chizhang} or utilize 2D projection of LiDAR \cite{nicolai2016deep,velas2018cnn,wang2019deeppco,li2019net,li2020dmlo} and the LiDAR odometry in 3D point clouds is underexplored. This paper aims to estimate the LiDAR odometry directly through raw 3D point clouds.

\begin{figure}[t]
	\centering
	\resizebox{0.90\columnwidth}{!}
	{
		\includegraphics[scale=1.0]{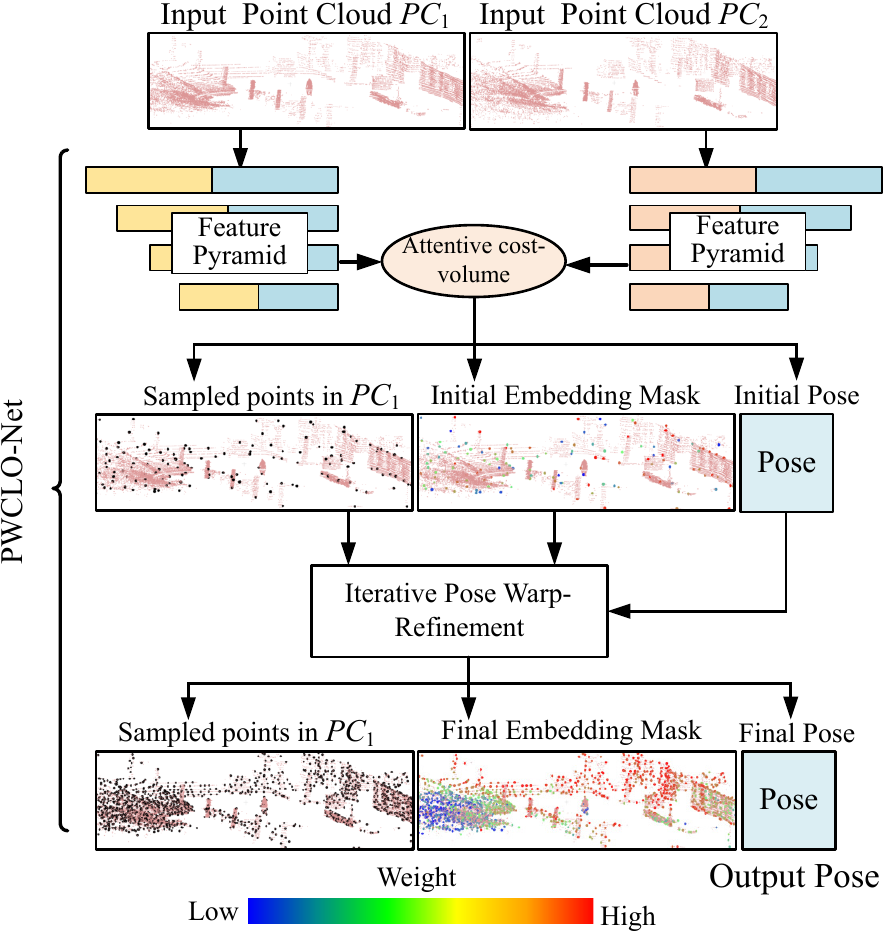}}
	\vspace{0mm}
	\caption{The Point Feature Pyramid, Pose Warping, and Attentive Cost Volume (PWC) structure in our proposed PWCLO-Net. The pose is refined layer by layer through iterative pose warp-refinement. The whole process is realized end-to-end by making all modules differentiable. In the LiDAR point clouds, the small red points are the whole point cloud. The big black points are the sampled points in $PC_1$. Distinctive colors of big points in embedding mask measure the contribution of sampled points to the pose estimation.}
	\label{fig:network_little}
	\vspace{-12pt}
\end{figure}

For the LiDAR odometry in 3D point clouds, there are three challenges: 1) As the discrete LiDAR point data are obtained in two consecutive frames separately, it is hard to find an precisely corresponding point pair between two frames; 2) Some points belonging to an object in a frame may not be seen in the other view if they are occluded by other objects or are not captured because of the limitation of LiDAR resolution; 3) Some points belonging to dynamic objects are not suitable to be used for the pose estimation because these points have uncertainty motions of the dynamic objects.

For the first challenge, Zheng et al. \cite{zheng2020lodonet} use matched keypoint pairs judged in 2D depth images. However, the correspondence is rough because of the discrete perception of LiDAR. The cost volume for 3D point cloud \cite{wu2019pointpwc,wang2020hierarchical} is adopted in this paper to obtain a weighted soft correspondence between two consecutive frames. For the second and third challenges, the mismatched points or dynamic objects, which do not conform to the overall pose, need to be filtered. LO-Net~\cite{li2019net} trains an extra mask estimation network~\cite{zhou2017unsupervised, yang2018unsupervised} by weighting the consistency error of the normal of 3D points. In our network, an internal trainable embedding mask is proposed to weigh local motion patterns from the cost volume to regress the overall pose. In this way, the mask can be optimized for more accurate pose estimation rather than depending on geometry correspondence. In addition, the PWC structure is built to capture the large motion in the layer of sparse points and refine the estimated pose in dense layers. As shown in Fig.~\ref{fig:network_little}, the embedding mask is also optimized hierarchically to obtain more accurate filtering information to refine pose estimation.

Overall, our contributions are as follows:
\vspace{-0.1cm}
\begin{itemize}
	\vspace{-0.2cm}
	\item The Point Feature Pyramid, Pose Warping, and Cost Volume (PWC) structure for the 3D LiDAR odometry task is built to capture the large motion between two frames and accomplish the trainable iterative 3D feature matching and pose regression.
	\vspace{-0.3cm}
	\item In this structure, the hierarchical embedding mask is proposed to filter mismatched points and convert the cost volume embedded in points to the overall ego-motion in each refinement level. Meanwhile, the embedding mask is optimized and refined hierarchically to obtain more accurate filtering information for pose refinement following the density of points.
	\vspace{-0.3cm}
    \item Based on the characteristic of the pose transformation,
    the pose warping and pose refinement are proposed to refine the estimated pose layer by layer iteratively. A totally end-to-end framework, named PWCLO-Net, is established, in which all modules are fully differentiable so that each process is no longer independent and combinedly optimized.
	\vspace{-0.3cm}
	\item Finally, our method is demonstrated on KITTI odometry dataset \cite{geiger2012we,geiger2013vision}. The evaluation experiments and ablation studies show the superiority of the proposed method and the effectiveness of each design. To the best of our knowledge, our method outperforms all recent learning-based LiDAR odometry and even outperforms the geometry-based LOAM with mapping optimization \cite{zhang2017low} on most sequences. 
	\vspace{-0.2cm}
\end{itemize}

\section{Related Work}
\vspace{-4pt}
\subsection{Deep LiDAR Odometry}
\vspace{-4pt}
Deep learning has gained impressive progress in visual odometry \cite{min2020voldor,yang2020d3vo}. However, the 3D LiDAR odometry with deep learning is still a challenging problem. In the beginning, Nicolai et al. \cite{nicolai2016deep} project two consecutive LiDAR point clouds to the 2D plane to obtain two 2D depth images, and then use the 2D convolution and fully connected (FC) layers to realize learning-based LiDAR odometry. Their work verified that the learning-based method is serviceable for the LiDAR odometry although their experiment results are not superior. Velas et al. \cite{velas2018cnn} also project LiDAR points to the 2D plane but use three channels to encode the information, including height, range, and intensity. Then the convolution and FC layers are used for the pose regression. The performance is superb when only estimating the translation but is poor when estimating the 6-DOF pose. Wang et al. \cite{wang2019deeppco} project point clouds to panoramic depth images and stack two frames together as input. Then the translation sub-network and FlowNet \cite{dosovitskiy2015flownet} orientation sub-network are used to estimate the translation and orientation respectively. \cite{li2019net} also preprocess 3D LiDAR points to 2D information but use the cylindrical projection \cite{chen2017multi}. Then, the normal of each 3D point is estimated to build consistency constraint between frames, and an uncertainty mask is estimated to mask dynamic regions. Zheng et al. \cite{zheng2020lodonet} extract matched keypoint pairs by classic detecting and matching algorithm in 2D spherical depth images projected from 3D LiDAR point clouds. Then the PointNet \cite{qi2017pointnet} based structure is used to regress the pose from the matched keypoint pairs. \cite{li2020dmlo} proposes a learning-based network to generate matching pairs with high confidence, then applies Singular Value Decomposition (SVD) to get the 6-DoF pose. \cite{cho2020} introduces an unsupervised learning method on LiDAR odometry.

\begin{figure*}[t]
	\centering
	\resizebox{1.00\textwidth}{!}
	{
	\includegraphics[scale=1.00]{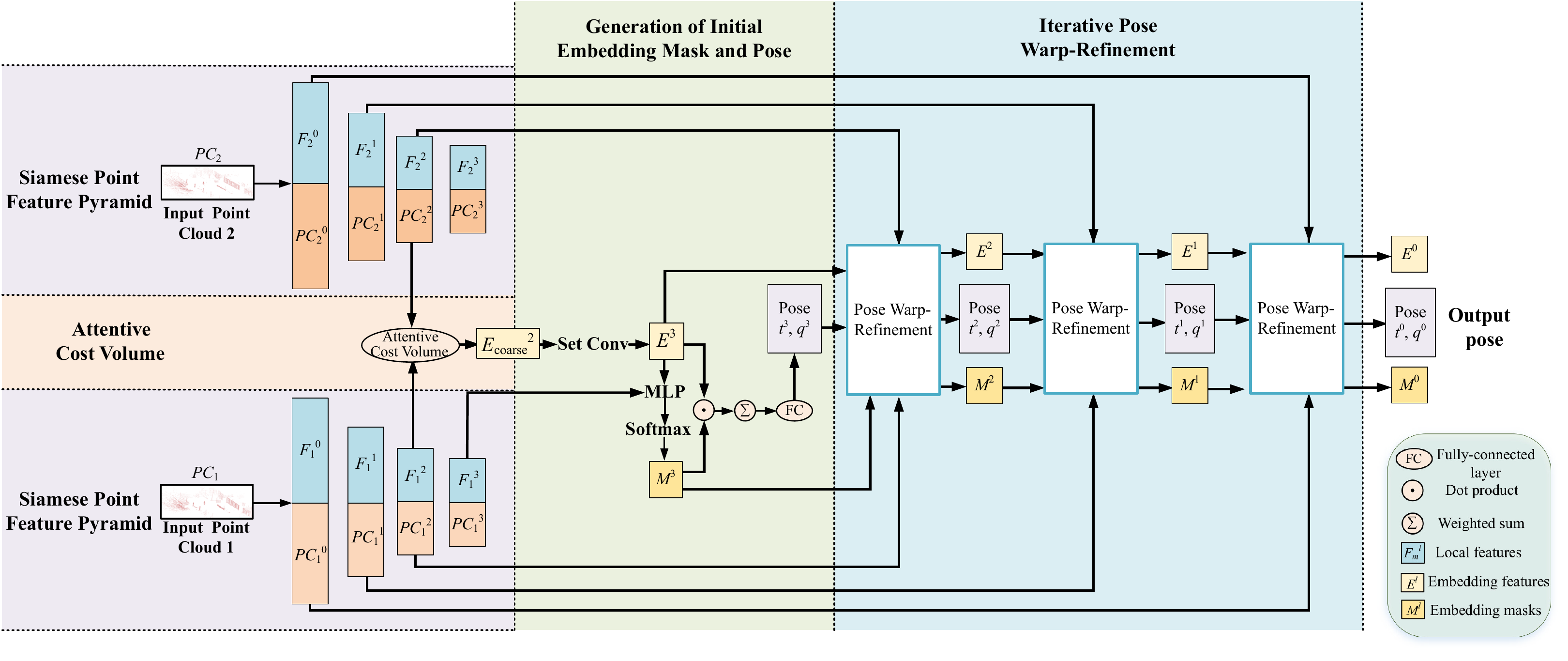}}
	\vspace{-7mm}
	\caption{The details of proposed PWCLO-Net architecture. The network is composed of four set conv layers in point feature pyramid, one attentive cost volume, one initial embedding mask and pose generation module, and three pose warp-refinement modules. The network outputs the predicted poses from four levels for supervised training. }
	\label{fig:network}
	\vspace{-12pt}
\end{figure*}

\vspace{-0.1cm}
\subsection{Deep Point Correlation}
\vspace{-4pt}
The above studies all use the 2D projection information for the LiDAR odometry learning, which converts the LiDAR odometry to the 2D learning problem. Wang et al. \cite{wang2019deeppco} compared the 3D point input and 2D projection information input based on the same 2D convolution model. It is found that the 3D input based method has a poor performance. As the development of 3D deep learning \cite{qi2017pointnet,qi2017pointnet++}, FlowNet3D \cite{liu2019flownet3d} proposes an embedding layer to learn the correlation between the points in two consecutive frames. After that, Wu et al. \cite{wu2019pointpwc} propose the cost volume method on point clouds, and Wang et al. \cite{wang2020hierarchical} develop the attentive cost volume method. The point cost volume involves the motion patterns of each point. It becomes a new direction and challenge to regress pose from the cost volume, and meanwhile, not all point motions are for the overall pose motion. We exploit to estimate the pose directly from raw 3D point cloud data and deal with the new challenges encountered.

What is more, we are inspired by the Pyramid, Warping, and Cost volume (PWC) structure in the flow network proposed by Su et al. \cite{sun2018pwc}. The work uses the three modules (Pyramid, Warping, and Cost volume) to refine the optical flow through a coarse-to-fine approach. The works \cite{wu2019pointpwc, wang2020hierarchical} on 3D scene flow also use the PWC structure to refine the estimated 3D scene flow in point clouds. In this paper, the idea is applied to the pose estimation refinement, and a PWC structure for LiDAR odometry is built for the first time.


\vspace{-0.2cm}
\section{PWCLO-Net}
\vspace{-4pt}
Our method learns the LiDAR odometry from raw 3D point clouds in an end-to-end approach with no need to pre-project the point cloud into 2D data, which is significantly different from the deep LiDAR odometry methods introduced in the related work section. 

The Fig.~\ref{fig:network} illustrates the overall structure of our PWCLO-Net. The inputs to the network are two point clouds $PC_1 = \{ {x_i}|{x_i} \in {\mathbb{R}^3}\} _{i = 1}^N$ and $PC_2 = \{ {y_j}|{y_j} \in {\mathbb{R}^3}\} _{j = 1}^N$ respectively sampled from two adjacent frames. These two point clouds are firstly encoded by the siamese feature pyramid consisting of several set conv layers as introduced in Sec.~\ref{sec:pyramid}. Then the attentive cost volume is used to generate embedding features, which will be described in Sec.~\ref{sec:cost}. To regress pose transformation from the embedding features, hierarchical embedding mask optimization is proposed in Sec.~\ref{sec:mask}. Next, the pose warp-refinement method is proposed in Sec.~\ref{sec:refine} to refine the pose estimation in a coarse-to-fine approach. Finally, the network outputs the quaternion $q \in {\mathbb{R}^4}$ and translation vector $t \in {\mathbb{R}^3}$.


\subsection{Siamese Point Feature Pyramid}\label{sec:pyramid}
\vspace{-4pt}
The input point clouds are usually disorganized and sparse in a large 3D space. A siamese feature pyramid consisting of several set conv layers is built to encode and extract the hierarchical features of each point cloud. Farthest Point Sampling (FPS)~\cite{qi2017pointnet++} and shared Multi-Layer Perceptron (MLP) are used. The formula of set conv is:
{\setlength\abovedisplayskip{4pt}\setlength\belowdisplayskip{4pt}\begin{equation} {f_i} = \mathop {MAX}\limits_{k = 1,2,...K} (MLP((x_i^k - x_i) \oplus f_i^k \oplus f_i^c)),\end{equation}}where $x_i$ is the obtained $i$-th sampled point by FPS. And $K$ points  $x_i^k$ $(k=1,2,...,K)$ are selected by $K$ Nearest Neighbors (KNN) around $x_i$. $f_i^c$ and $f_i^k$ are the local features of $x_i$ and $x_i^k$ (they are null for the first layer in the pyramid). ${f_i}$ is the output feature located at the central point $x_i$. $\oplus$ denotes the concatenation of two vectors, and $\mathop {MAX}\limits_{k = 1,2,...K}() $ indicates the max pooling operation. The hierarchical feature pyramid is constructed as shown in Fig.~\ref{fig:network}. The siamese pyramid \cite{chopra2005learning} means that the learning parameters of the built pyramid are shared for these two point clouds.

\subsection{Attentive Cost Volume}\label{sec:cost}
\vspace{-4pt}
Next, the point cost volume with attention in \cite{wang2020hierarchical} is adopted here to associate two point clouds. The cost volume generates point embedding features by associating two point clouds after the feature pyramid. 

\begin{figure}[t]
	\centering
	\resizebox{1.0\columnwidth}{!}
	{
		\includegraphics[scale=1.00]{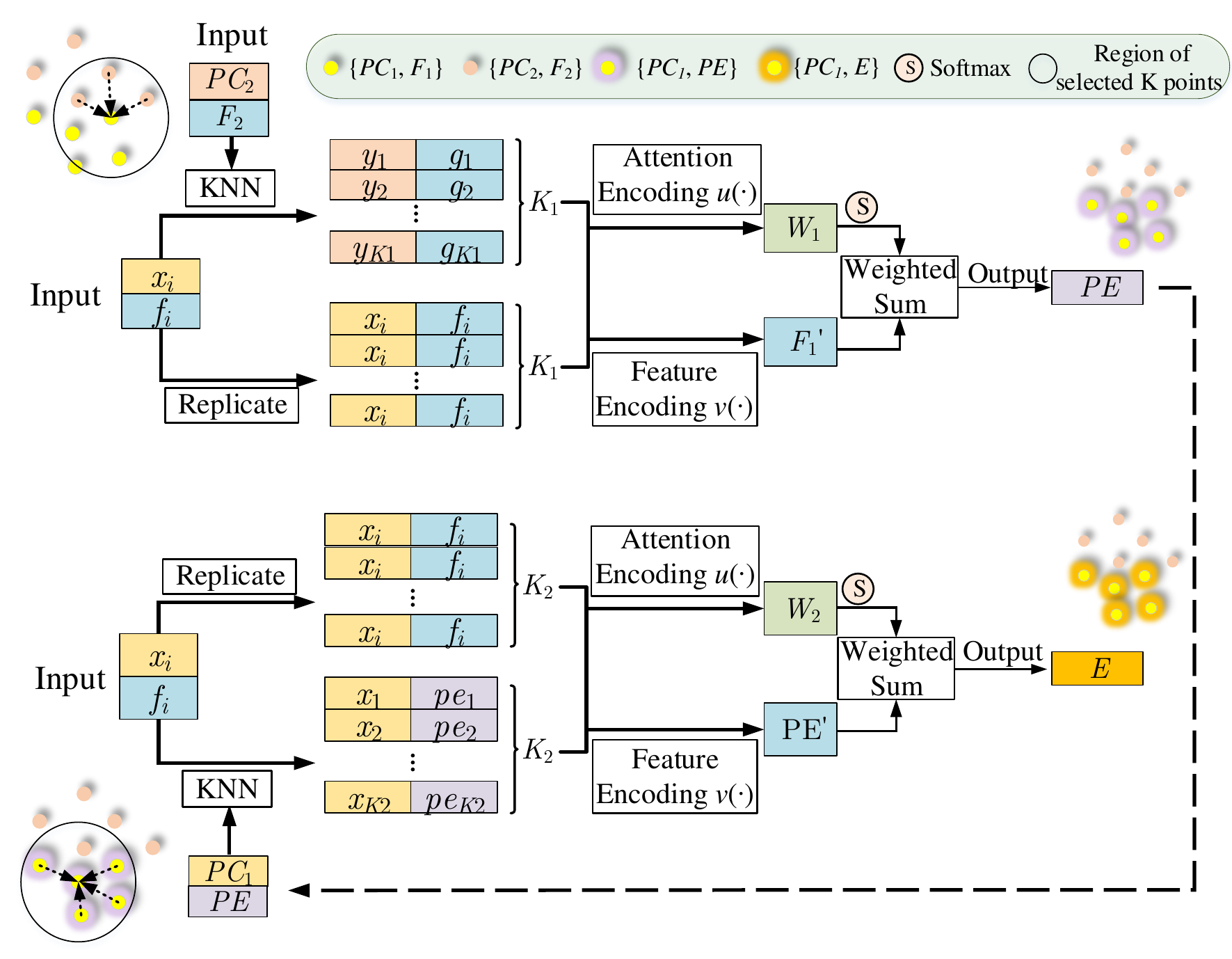}}
	\vspace{-5mm}
	\caption{Attention Cost-volume. This module takes two frames of point clouds with their local features as input and associates the two point clouds. Finally, the module outputs the embedding features located in $PC_1$.}
	\label{fig:cost}
	\vspace{-8pt}
\end{figure}

The embedding features contain the point correlation information between two point clouds. As shown in Fig.~\ref{fig:cost},  $F_1 = \{ {f_i}|{f_i} \in {\mathbb{R}^c}\} _{i = 1}^n$ is the features of point cloud $PC_1 = \{ {x_i}|{x_i} \in {\mathbb{R}^3}\} _{i = 1}^n$ and $F_2 = \{ {g_j}|{g_j} \in {\mathbb{R}^c}\} _{j = 1}^n$ is the features of point cloud $PC_2 = \{ {y_j}|{y_i} \in {\mathbb{R}^3}\} _{i = 1}^n$. The embedding features between two point clouds are calculated as  follows:
\begin{equation}w_{1, i}^k = softmax(u(x_i,y_j^k,f_i,g_j^k))_{k = 1}^{K_1},\end{equation}\vspace{-0.5cm}
\begin{equation}pe_{i} = \sum\limits_{k = 1}^{k_1} {{w_{1,i}^k} \odot v(x_i,y_j^k,f_i,g_j^k)},\end{equation}\vspace{-0.2cm}
\begin{equation}w_{2,i}^k = softmax(u(x_i,x_i^k,pe_{i},pe_{i}^k))_{k = 1}^{K_2},\end{equation}\vspace{-0.2cm}
\begin{equation}e_{i} = \sum\limits_{k = 1}^{k_2} {w_{2,i}^k \odot v(x_i,x_i^k,pe_{i},pe_{i}^k)},\end{equation}
where $y_j^k$ and $g_j^k$ represent the coordinates and local features of the selected $K_1$ points in $PC_2$ respectively.  $\odot$ represents dot product. $u(\cdot)$ and $v(\cdot )$ represent the attention encoding and feature encoding functions referring to \cite{wang2020hierarchical}. $u(\cdot)$ encodes the 3D Euclidean space information and point features to generate attention weights, and $v(\cdot)$ represents the further feature encoding on spacial information and features of two frames of point cloud. The output $E = \{ {e_{i}}|{e_{i}} \in {\mathbb{R}^c}\} _{i = 1}^n$ is the embeding features in $PC_1$.

\subsection{Hierarchical Embedding Mask Optimization}\label{sec:mask}
\vspace{-4pt}
It is a new problem to convert the embedding features $E$ to a global consistent pose transformation between two frames. In this subsection, a novel embedding mask is proposed to generate pose transformation from embedding features.

It should be noted that some points may belong to dynamic objects or be occluded in the other frame. It is necessary to filter out these points and keep the points that are of value to the LiDAR odometry task. To deal with this, the embedding features $E = \{ {e_{i}}|{e_{i}} \in {\mathbb{R}^c}\} _{i = 1}^n$ and the features $F_1$ of $PC_1$ are input to a shared MLP followed by the softmax operation along the point dimension to obtain the embedding mask (as the initial embedding mask in Fig.~\ref{fig:network}):
\vspace{-0.2cm}
\begin{equation}
{M} = softmax(sharedMLP({E} \oplus F_1)),\vspace{-0.2cm}\end{equation}
where $M = \{ {m_{i}}|{m_{i}} \in {\mathbb{R}^c}\} _{i = 1}^n$ represents trainable masks for prioritizing embedding features of $n$ points in $PC_1$. Each point has a characteristic weight ranging from 0 to 1. The smaller the weight of a point is, the more likely the point needs to be filtered out, and vice versa. Then, the quaternion  $q \in {\mathbb{R}^4}$ and translation vector $t \in {\mathbb{R}^3}$ can be generated by weighting embedding features and FC layers separately, and $q$ is normalized to accord with the characteristic of rotation.\vspace{-0.3cm}
\begin{equation}
q = \frac{{FC(\sum\limits_{i = 1}^n {{e_i} \odot {m_i}} )}}{{\left| {FC(\sum\limits_{i = 1}^n {{e_i} \odot {m_i}} )} \right|}}\label{eqn:q},\vspace{-0.2cm}\end{equation}
\vspace{-0.1cm}
\begin{equation}
t = FC(\sum\limits_{i = 1}^n {{e_{i}} \odot {m_{i}}} )\label{eqn:t}.\vspace{-0.2cm}\end{equation}

The trainable mask $M$ is also part of the hierarchical refinement. As shown in Fig.~\ref{fig:network}, the embedding mask is propagated to denser layers of the point cloud just like embedding features $E$ and pose. The embedding mask is also optimized in a coarse-to-fine approach during the warp-refinement process, making the final mask estimation and the calculation of pose transformation accurate and reliable. We call this process the hierarchical embedding mask optimization.

\subsection{Pose Warp-Refinement Module}\label{sec:refine}
\begin{figure}[t]
	\centering
	\vspace{-0mm}
	\resizebox{1.00\columnwidth}{!}
	{
		\includegraphics[scale=1.00]{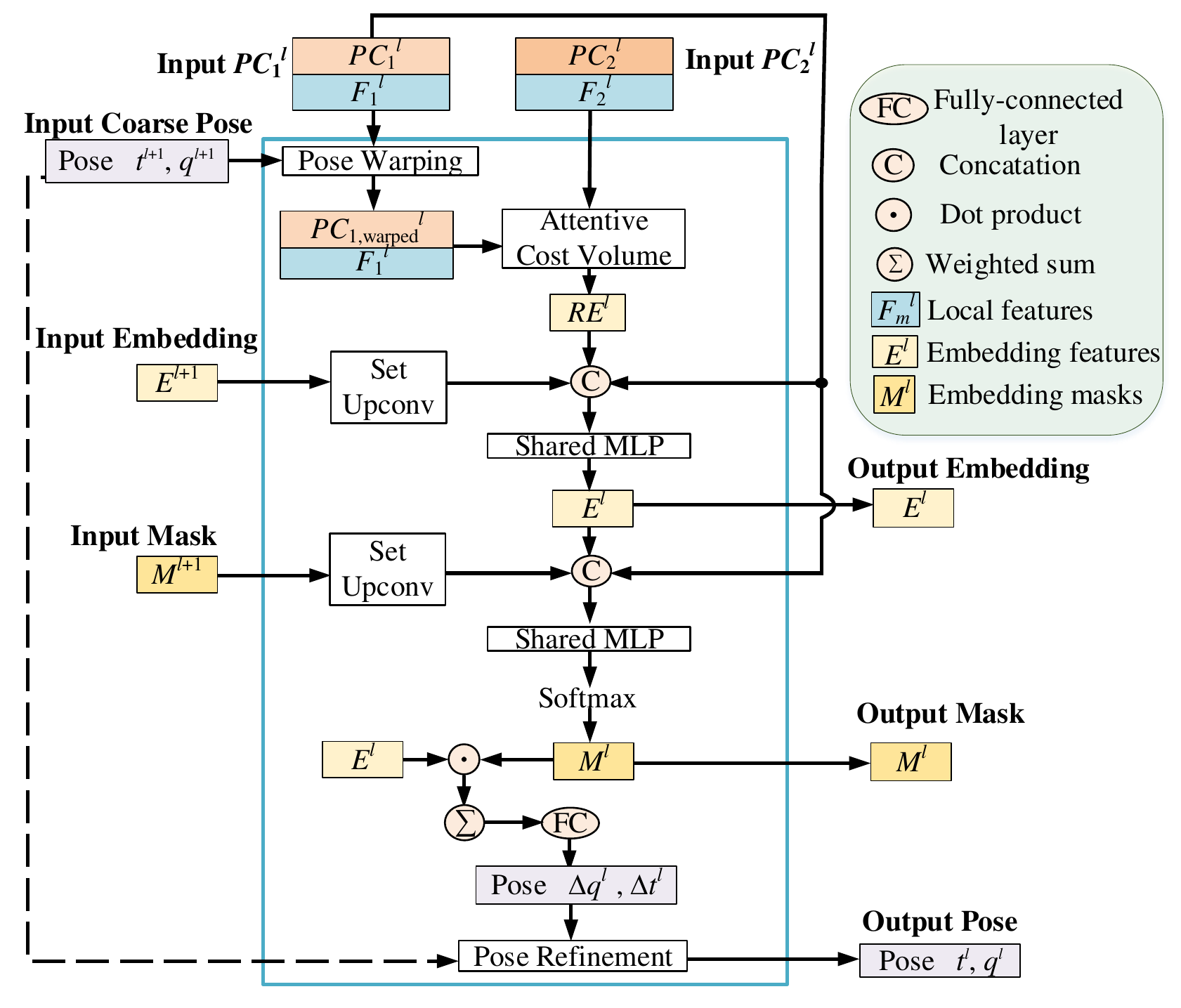}}
	\vspace{-5mm}
	\caption{The details of the proposed Pose Warp-Refinement module at the $l$-th level.}
	\label{fig:warping}
	\vspace{-10pt}
\end{figure}

\vspace{-4pt}
To achieve the coarse-to-fine refinement process in an end-to-end fashion, we propose the differentiable warp-refinement module based on pose transformation as shown in Fig.~\ref{fig:warping}. This module contains several key parts: set upconv layer, pose warping, embedding feature and embedding mask refinement, and pose refinement.

\vspace{5pt}
\noindent{}{\bf Set Upconv Layer:}
To refine the pose estimation in a coarse-to-fine approach, the set upconv layer \cite{liu2019flownet3d} is adopted here to enable features of point cloud to propagate from sparse level to dense level. Embedding features $E^{l+1}$ and embedding masks $M^{l+1}$ of $l+1$ layer are propagated through set upconv layer to obtain coarse embedding features $CE^{l}=\{ {ce_{i}^l}|{ce_{i}^l} \in {\mathbb{R}^{c^l}}\} _{i = 1}^{n^l}$ and coarse embedding masks $CM^{l}= \{ {cm_{i}^l}|{cm_{i}^l} \in {\mathbb{R}^{c^l}}\} _{i = 1}^{n^l}$ that need to be optimized at the $l$-th level.

\vspace{5pt}
\noindent{}{\bf Pose Warping:}
The process of pose warping means that the quaternion $q^{l+1}$ and translation vector $t^{l+1}$ from the ${(l+1)}$-th level are applied to warp $PC_1^l=\{ {x_{i}^l}|{x_{i}^l} \in {\mathbb{R}^{c^l}}\} _{i = 1}^{n^l}$ to generate $PC_{1,warped}^l=\{ {x_{i,warped}^l}|{x_{i,warped}^l} \in {\mathbb{R}^{c^l}}\} _{i = 1}^{n^l}$. The warped $PC_{1,warped}^l$ is closer to $PC_2^l$ than original $PC_{1}^l$, which makes the residual motion estimation easier at the $l$-th level. The equation of warping transformation is as follows:\vspace{-0.2cm}
\begin{equation}
[0,x_{i,warped}^l] = {q^{l+1}}[0,x_{i}^l](q^{l+1})^{-1}+[0, t^{l+1}].\vspace{-0.2cm}\end{equation}

Then, the attentive cost volume between $PC{_{1,warped}^l}$ and $PC_2^l$ is re-calculated to estimate the residual motion. Following the approach introduced in Sec.~\ref{sec:cost}, re-embedding features between $PC{_{1,warped}^l}$ and $PC_2^l$ are calculated and denoted as $RE^l = \{ {re_{i}^l}|{re_{i}^l} \in {\mathbb{R}^{c^l}}\} _{i = 1}^{n^l}$.  

\vspace{5pt}
\noindent{}{\bf Embedding Feature and Embedding Mask Refinement:}
The generated coarse embedding feature $ce_i^{l}$, the re-embedding feature $re_{i}^l$, and the features $f_i^l$ of $PC_1^l$ are concatenated and input to a shared MLP to obtain embedding features $E^{l}=\{ {e_{i}^l}|{e_{i}^l} \in {\mathbb{R}^{c^l}}\} _{i = 1}^{n^l}$ at the $l$-th level:
\vspace{-0.2cm}
\begin{equation}
	\vspace{-0.1cm}
e_{i}^l = MLP(ce_{i}^l \oplus re_{i}^{l}  \oplus f_i^l).
\vspace{-0.2cm}\end{equation}The output of this MLP is the optimized embedding features of $l$-th level, which will not only participate in the following pose generation operation but also be output as the input to the next level warp-refinement module. 

Like the refinement of the embedding feature, the newly generated embedding feature $e_{i}^l$, the generated coarse embedding mask $cm_i^{l}$, and the local feature $f_i^l$ of $PC_1^l$ are concatenated and input to a shared MLP and softmax operation along the point dimension to obtain the embedding mask $M^{l}= \{ {m_{i}^l}|{m_{i}^l} \in {\mathbb{R}^{c^l}}\} _{i = 1}^{n^l}$ at the $l$-th level:
\vspace{-0.1cm}
\begin{equation}
M^l = softmax(sharedMLP(E^l \oplus CM^{l}  \oplus F^l)).
\vspace{-0.2cm}
\end{equation}

\vspace{0pt}
\noindent{}{\bf Pose Refinement:}
The residual $\Delta q^l$ and $\Delta t^l$ can be obtained from the refined embedding features and mask following the formulas~(\ref{eqn:q}) and (\ref{eqn:t}) in Sec.~\ref{sec:mask}. Lastly, the refined quaternion $q^l$ and translation vector $t^l$ of the $l$-th level can be calculated by:\vspace{-0.2cm}\begin{equation}{q^l} = {\Delta q^{l}}{q^{l + 1}},\end{equation}\vspace{-0.5cm}\begin{equation}[0,t^l] = {\Delta q^{l}}[0,t^{l+1}](\Delta q^{l})^{-1}+[0, \Delta t^{l}].\end{equation}

\subsection{Training Loss}
\vspace{-4pt}
The network outputs quaternion $q^l$ and translation vector $t^l$ from four different levels of point clouds. The outputs of each level will enter into a designed loss function and be used to calculate the supervised loss $\ell^l$. Due to the difference in scale and units between translation vector $t$ and quaternion $q$, two learnable parameters $s_x$ and $s_q$ are introduced like previous deep odometry work \cite{li2019net}. The training loss function at the $l$-th level is: \begin{equation}\begin{gathered}
{\ell ^l} = {\left\| {{t_{gt}} - {t^l}} \right\|}exp( - {s_x}) + {s_x} \hfill \\
{\text{   }} + {\left\| {{q_{gt}} - \frac{{{q^l}}}{{\left\| {{q^l}} \right\|}}} \right\|_2}exp( - {s_q}) + {s_q},  
\end{gathered} \label{eqn:sxsq} \end{equation}
where $\left\|  \cdot  \right\|$ and ${\left\|  \cdot  \right\|_2}$ denotes the ${\ell _1}$-norm and the ${\ell _2}$-norm respectively. ${{t_{gt}}}$ and ${{q_{gt}}}$ are the ground-truth translation vector and quaternion respectively generated by the ground-truth pose transformation matrix. Then, a multi-scale supervised approach is adopted. The total training loss $\ell$ is:\vspace{-0.2cm}
\begin{equation}\ell  = \sum\nolimits_{l = 1}^L {{\alpha ^l}} {\ell ^l}\label{eqn:l},\end{equation}
where $L$ is the total number of warp-refinement levels and ${{\alpha ^l}}$ denotes the weight of the $l$-th level.

\setlength{\tabcolsep}{0.9mm}
\begin{table*}[t]
	\centering
	\footnotesize
	\begin{center}
		\resizebox{1.0\textwidth}{!}
		{
			\begin{tabular}{l||cc|cc|cc|cc|cc|cc|cc|cc|cc|cc|cc||cc}
				\toprule
				&  \multicolumn{2}{c|}{00$^*$}  &\multicolumn{2}{c|}{01$^*$}      & \multicolumn{2}{c|}{02$^*$} & \multicolumn{2}{c|}{03$^*$} &  \multicolumn{2}{c|}{04$^*$} & \multicolumn{2}{c|}{05$^*$} & \multicolumn{2}{c|}{06$^*$} & \multicolumn{2}{c|}{07} & \multicolumn{2}{c|}{08} & \multicolumn{2}{c|}{09} &\multicolumn{2}{c||}{10} &\multicolumn{2}{c}{Mean on 07-10} \\ 
				\cline{2-25}\noalign{\smallskip}
				
				\multirow{-2}{*}{\begin{tabular}[c]{@{}c@{}}Method \end{tabular}}
				&  $t_{rel}$  & $r_{rel}$   & $t_{rel}$                       & $r_{rel}$               & $t_{rel}$                          & $r_{rel}$   & $t_{rel}$ & $r_{rel}$   & $t_{rel}$                          & $r_{rel}$   & $t_{rel}$ & $r_{rel}$    & $t_{rel}$                          & $r_{rel}$   & $t_{rel}$ & $r_{rel}$ & $t_{rel}$                          & $r_{rel}$   & $t_{rel}$ & $r_{rel}$    & $t_{rel}$ & $r_{rel}$  & $t_{rel}$ & $r_{rel}$       \\
				\hline\hline
				\noalign{\smallskip}

				Full LOAM \cite{zhang2017low}    
                &1.10 &	0.53 
                &2.79 &	0.55 
                &1.54 & 0.55 
                &1.13 &	0.65 
                &1.45 & 0.50 
                &0.75 &	0.38 
                &0.72 & 0.39 
                &0.69 & 0.50 
                &1.18 & 0.44 
                &1.20 & 0.48 
                &1.51 & \bf0.57 
                & 1.145 &  0.498
                \\
				\hline 
				ICP-po2po    
				&6.88&	2.99
				&11.21&	2.58
				& 8.21	&3.39
				&11.07&	5.05
				& 6.64	&4.02
				& 3.97&	1.93
				&1.95	&1.59
				&5.17	&3.35
				&10.04	&4.93
				&6.93	&2.89
				&8.91	&4.74
				&7.763	&3.978
				\\ 
				
				ICP-po2pl     
				&3.80 &	1.73
				&13.53 & 2.58 
				&9.00 & 2.74 
				&2.72 & 1.63 
				&2.96 & 2.58 
				&2.29 & 1.08 
				&1.77 & 1.00 
				&1.55 & 1.42 
				&4.42 & 2.14 
				&3.95 & 1.71 
				&6.13 & 2.60 
				&4.013 & 1.968
				\\

				GICP \cite{segal2009generalized}    
				&1.29 & 0.64 
				&4.39 & 0.91 
				&2.53 & 0.77 
				&1.68 & 1.08 
				&3.76 & 1.07 
				&1.02 & 0.54 
				&0.92 & 0.46 
				&0.64 & 0.45 
				&1.58 & 0.75 
				&1.97 & 0.77 
				&1.31 &0.62 
				&1.375 & 0.648
				\\ 
				CLS \cite{velas2016collar}    
				&2.11 & 0.95 
				&4.22 & 1.05 
				&2.29 & 0.86
				&1.63 & 1.09 
				&1.59 & 0.71 
				&1.98 & 0.92 
				&0.92 & 0.46 
				&1.04 & 0.73 
				&2.14 & 1.05 
				&1.95 & 0.92 
				&3.46 & 1.28 
				&2.148 &  0.995
				\\ 
				
				Velas et al. \cite{velas2018cnn}    
				&3.02 &	NA
				&4.44 &	NA
				&3.42 & NA
				&4.94 &	NA
				&1.77 & NA
				&2.35 &	NA
				&1.88 & NA
				&1.77 & NA
				&2.89 & NA
				&4.94 & NA
				&3.27 & NA
				&3.218 & NA

				\\ 
				LO-Net \cite{li2019net}    
				&1.47 & 0.72 
				&1.36 & 0.47 
				&1.52 & 0.71 
				&1.03 & 0.66 
				&0.51 & 0.65 
				&1.04 & 0.69 
				&0.71 & 0.50 
				&1.70 & 0.89 
				&2.12 & 0.77 
				&1.37 & 0.58 
				&1.80 & 0.93 
				&1.748 & 0.793
				\\ 
				DMLO \cite{li2020dmlo}    
				& NA   & NA
				& NA   & NA
				& NA   & NA
				& NA   & NA
				& NA   & NA
				& NA   & NA
				& NA   & NA
				&0.73 & 0.48
				&\bf 1.08 & \bf 0.42
				&1.10 & 0.61
				&\bf 1.12 & 0.64
				&\bf 1.008 & 0.538
				\\
				LOAM w/o mapping    
				&15.99  &6.25	  
				& 3.43 &0.93	 
				& 9.40 & 3.68 
				& 18.18 &9.91	  
				& 9.59 &  4.57
				& 9.16 &  4.10
				& 8.91 &  4.63
				& 10.87 &  6.76
				& 12.72 &  5.77 
				& 8.10 &  4.30
				& 12.67 &  8.79
				& 11.090 & 6.405
				\\             					
				Ours      
				&\bf0.78 & \bf0.42
				&\bf0.67 & \bf0.23
				&\bf0.86 & \bf0.41
				&\bf0.76 & \bf0.44
				&\bf0.37 & \bf0.40
				&\bf0.45 & \bf0.27
				&\bf0.27 & \bf0.22
				&\bf0.60 & \bf0.44
				&1.26 &  0.55
				&\bf0.79 & \bf0.35
				&1.69 & 0.62
				&1.085  &\bf 0.490    
				\\ \bottomrule
			\end{tabular}
		}
	\end{center}
    \vspace{-8pt}
	\caption{The LiDAR odometry experiment results on KITTI odometry dataset \cite{geiger2013vision}. $t_{rel}$ and $r_{rel}$ mean the average translational RMSE (\%) and rotational RMSE ($^{\circ}$/100m) respectively on all possible subsequences in the length of $100,200,...,800m$. `$^*$' means the training sequence. LOAM is a complete SLAM system, including back-end optimization and others only include odometry. The data other than the last three lines are from \cite{li2019net}. The results of LOAM w/o mapping are obtained by running their published codes. The best results are bold.}
	\vspace{-8pt}		
	\label{table:lidar}
\end{table*}

\setlength{\tabcolsep}{0.9mm}
\begin{table*}[t]
	\centering
	\footnotesize
	\begin{center}
		\resizebox{0.9\textwidth}{!}
		{
			\begin{tabular}{l||cc|cc|cc|cc|cc|cc|cc|cc|cc|cc|cc||cc}
				\toprule
				&  \multicolumn{2}{c|}{00$^*$}  &\multicolumn{2}{c|}{01$^*$}      & \multicolumn{2}{c|}{02$^*$} & \multicolumn{2}{c|}{03$^*$} &  \multicolumn{2}{c|}{04$^*$} & \multicolumn{2}{c|}{05$^*$} & \multicolumn{2}{c|}{06$^*$} & \multicolumn{2}{c|}{07} & \multicolumn{2}{c|}{08} & \multicolumn{2}{c|}{09$^*$} &\multicolumn{2}{c||}{10$^*$} &\multicolumn{2}{c}{Mean on 07-08} \\ 
				\cline{2-25}\noalign{\smallskip}
				
				\multirow{-2}{*}{\begin{tabular}[c]{@{}c@{}}Method \end{tabular}}
				&  $t_{rel}$  & $r_{rel}$   & $t_{rel}$                       & $r_{rel}$               & $t_{rel}$                          & $r_{rel}$   & $t_{rel}$ & $r_{rel}$   & $t_{rel}$                          & $r_{rel}$   & $t_{rel}$ & $r_{rel}$    & $t_{rel}$                          & $r_{rel}$   & $t_{rel}$ & $r_{rel}$ & $t_{rel}$                          & $r_{rel}$   & $t_{rel}$ & $r_{rel}$    & $t_{rel}$ & $r_{rel}$  & $t_{rel}$ & $r_{rel}$       \\

				\hline\hline
				\noalign{\smallskip}
			
				LodoNet \cite{zheng2020lodonet}      
				&1.43	&0.69
				&0.96	&0.28
				&1.46	&0.57
				&2.12	&0.98
				&0.65	&0.45
				&1.07	&0.59
				&0.62	&0.34
				&1.86	&1.64
				&2.04	&0.97
				&0.63	&0.35
				&1.18	&0.45
				&1.950	&1.305
				
				\\                		
				
				Ours      
				&\bf0.75 & \bf0.36
				&\bf0.57 & \bf0.21
				&\bf0.83 & \bf0.37
				&\bf0.90 & \bf0.38
				&\bf0.45 & \bf0.39
				&\bf0.53 & \bf0.31
				&\bf0.37 & \bf0.22
				&\bf0.61 & \bf0.43
				&\bf1.29 & \bf0.57
				&\bf0.55 & \bf0.24
				&\bf0.61 &\bf0.39
				&\bf0.950 &\bf0.500
				\\ \bottomrule
			\end{tabular}
		}
	\end{center}
    \vspace{-8pt}
	\caption{The LiDAR odometry experiment results on KITTI odometry dataset \cite{geiger2013vision} compared with \cite{zheng2020lodonet}. As \cite{zheng2020lodonet} is trained on sequences 00-06, 09-10  and tested on sequences 07-08, we train and test our model like this to make comparisons with \cite{zheng2020lodonet}.}
	\vspace{-8pt}
	\label{table:lidar_7_8}
\end{table*}

\setlength{\tabcolsep}{1.3mm}
\begin{table}[t]
	\centering
	\footnotesize
	\begin{center}
		\resizebox{0.8\columnwidth}{!}
		{
			\begin{tabular}{l||cc|cc||cc}
				\toprule
				&    \multicolumn{2}{c|}{04} &\multicolumn{2}{c||}{10} &\multicolumn{2}{c}{Mean} \\ 
				\cline{2-7}\noalign{\smallskip}
				
				\multirow{-2}{*}{\begin{tabular}[c]{@{}c@{}}Method \end{tabular}}
				&  $t_{rel}$  & $r_{rel}$   & $t_{rel}$                       & $r_{rel}$               & $t_{rel}$                          & $r_{rel}$      \\
				\hline\hline
				\noalign{\smallskip}

				DeepPCO \cite{wang2019deeppco}    
				
				&2.63 & 3.05 
				
				&2.47 & 6.59 
				&2.550 & 4.820
				\\

				Ours      
				
				&\bf0.73 & \bf0.43
				
				&\bf1.57 & \bf0.57
				&\bf1.150    & \bf 0.500 
				\\ \bottomrule
			\end{tabular}
		}
	\end{center}
    \vspace{-8pt}
	\caption{The LiDAR odometry results on sequences 04 and 10 of KITTI odometry dataset \cite{geiger2013vision}. As \cite{wang2019deeppco} is trained on sequences 00-03, 05-09 and only reports testing results on sequences 04 and 10, we train and test our model like this to make comparisons with it.}
	\vspace{-8pt}		
	\label{table:lidar4_10}
	\vspace{0.2cm}
\end{table}

\vspace{-0.2cm}

\setlength{\tabcolsep}{1.3mm}
\begin{table}[t]
	\centering
	\footnotesize
	\begin{center}
		\resizebox{0.8\columnwidth}{!}
		{
			\begin{tabular}{l||cc|cc||cc}
				\toprule
				&    \multicolumn{2}{c|}{09} &\multicolumn{2}{c||}{10} &\multicolumn{2}{c}{Mean} \\ 
				\cline{2-7}\noalign{\smallskip}
				
				\multirow{-2}{*}{\begin{tabular}[c]{@{}c@{}}Method \end{tabular}}
				&  $t_{rel}$  & $r_{rel}$   & $t_{rel}$                       & $r_{rel}$               & $t_{rel}$                          & $r_{rel}$      \\
				\hline\hline
				\noalign{\smallskip}

				Cho et al. \cite{cho2020}    
				
				&4.87 & 1.95 
				
				&5.02 & 1.83 
				&4.945 & 1.890
				\\

				Ours      
				
				&\bf0.73 & \bf0.40
				
				&\bf1.16 & \bf0.78
				&\bf0.945    & \bf 0.295 
				\\ \bottomrule
			\end{tabular}
		}
	\end{center}
    \vspace{-8pt}
	\caption{The LiDAR odometry results on sequences 09 and 10 of KITTI odometry dataset \cite{geiger2013vision}. As \cite{cho2020} applies unsupervised training on sequences 00-08, and reports testing results on sequences 09 and 10, we train and test our model like this to make fair comparisons.}
	\vspace{-6pt}		
	\label{table:lidar09_10}
	\vspace{-0.2cm}
\end{table}

\section{Implementation}\label{sec:implementation}
\vspace{-4pt}

\subsection{KITTI Odometry Dataset}
\vspace{-4pt}
KITTI odometry dataset \cite{geiger2012we,geiger2013vision} is composed of 22 independent sequences. The Velodyne LiDAR point clouds in the dataset are used in our experiments. All scans of point clouds have XYZ coordinates and the reflectance information. Sequences 00-10 (23201 scans) contain ground truth pose (trajectory), while there is no ground truth publicly available for the remaining sequences 11-21 (20351 scans). By driving under different road environments, such as highways, residential roads, and campus roads, the sampling vehicle captures point clouds for the LiDAR odometry task from different environments.

\vspace{5pt}
\noindent{}{\bf Data Preprocessing:}
Only coordinates of LiDAR points are used in our method. Because the ground truth poses are denoted in the left camera coordinate system, all the training and evaluation processes of this network are carried out in the left camera coordinate system. Thus, the captured point clouds from the Velodyne LiDAR are firstly transformed to the left camera coordinate system by:
\vspace{-0.2cm}
\begin{equation}
{P_{cam}} = {T_r}P_{vel},
\vspace{-0.2cm}\end{equation}
where ${P_{cam}}$ and ${P_{vel}}$ are the point cloud coordinates in the left camera coordinate system and LiDAR coordinate system respectively, and ${T_r}$ is the calibration matrix of each sequence. 
Moreover, the point cloud collected by the LiDAR sensor often contains outliers at the edge of the point cloud in each frame. This is often because objects are far away from the LiDAR sensor, thus forming incomplete point clouds in the edge. To filter out these outlier points, the points out of the $30 \times 30m^2$ square area around the vehicle are filtered out for each point cloud $P_{cam}$. To speed up the data reading and training, the ground less than $0.55m$ in height is removed. For our model, the performances of removing and reserving the ground are similar. The detailed comparison is shown in the supplementary material.

\vspace{4pt}
\noindent{}{\bf Data Augmentation:}
We augment the training dataset by the augmentation matrix ${T_{aug}}$, generated by the rotation matrix ${R_{aug}}$ and translation vector ${t_{aug}}$.
Varied values of yaw-pitch-roll Euler angles are generated by Gaussian distribution around $0^{\circ}$. Then the ${R_{aug}}$ can be obtained from these random Euler angles. Similarly, the ${t_{aug}}$ is generated by the same process.
The composed ${T_{aug}}$ is then used to augment the $PC_1$ to obtain new point clouds $PC_{1,aug}$ by: \vspace{-0.2cm}
\begin{equation} \vspace{-0.2cm}
{PC_{1,aug}} = {T_{aug}}{PC_1}.\end{equation}

Correspondingly, the ground truth transformation matrix is also modified as:\vspace{-0.2cm}
\begin{equation}\vspace{-0.2cm}
{T_{trans}} = {T_{aug}}{T_p},\end{equation}
where ${T_p}$ denotes the original ground truth pose transformation matrix from $PC_1$ to $PC_2$.
${T_{trans}}$ is then used to generate ${q_{gt}}$ and ${t_{gt}}$ to supervise the training of the network.

\subsection{Network Details}
\vspace{-4pt}
In the training and evaluation process, the input $N$ points are randomly sampled from the point clouds of two frames separately. It is unnecessary that the original input two point clouds have the same number of points. And $N$ is set to be 8192 in the proposed network. Each layer in MLP contains the ReLU activation function, except for the FC layer. For shared MLP, $1 \times 1$ convolution with $1$ stride is the implement manner. The detailed layer parameters including each linear layer width in MLP are described in the supplementary material. All training and evaluation experiments are conducted on a single NVIDIA RTX 2080Ti GPU with TensorFlow 1.9.0. The Adam optimizer is adopted with ${\beta _1} = 0.9$, ${\beta _2} = 0.999$. The initial learning rate is 0.001 and exponentially decays every 200000 steps until 0.00001. The initial values of the trainable parameters $s_x$ and $s_q$ are set as 0.0 and -2.5 respectively in formula~(\ref{eqn:sxsq}). For formula~(\ref{eqn:l}), ${\alpha _1}$ = 1.6, ${\alpha _2}$ = 0.8, ${\alpha _3}$ = 0.4, and $L=4$. The batch size is 8.

\vspace{-0.1cm}

\section{Experimental Results}
\vspace{-4pt}
In this section, the quantitative and qualitative results of the network performance on the LiDAR odometry task are demonstrated and compared with state-of-the-art methods. Then, the ablation studies are presented. Finally, the embedding mask is visualized and discussed. 
\subsection{Performance Evaluation}
\vspace{-4pt}
As there are several modes to divide the training/testing sets for published papers \cite{li2019net,zheng2020lodonet,wang2019deeppco,zheng2020lodonet,cho2020}, in order to fairly compare with all current methods as we know, we train/test our model four times. 

 \noindent{}{\bf Using sequences 00-06/07-10 as training/testing sets:} Quantitative results are listed in Table~\ref{table:lidar}. ICP-point2point (ICP-po2po), ICP-point2plane (ICP-po2pl), GICP \cite{segal2009generalized}  , CLS \cite{velas2016collar} are several classic LiDAR odometry estimation methods based on Iterative Closest Point (ICP) \cite{besl1992method}. LOAM \cite{zhang2017low} is based on the matching of extracted line and plane features, which has a similar idea with our method, and it is a hand-crafted method. It achieves the best results among LiDAR-based methods in the KITTI odometry evaluation benchmark \cite{geiger2013vision}. Velas et al. \cite{velas2018cnn} is a learning-based method. It has a good performance when only the translation is estimated, but the performance will decrease when estimating the 6-DOF pose. LO-Net \cite{li2019net} and DMLO \cite{li2020dmlo} are learning-based LiDAR odometry methods that have comparable results, but they have no codes publicly available, so we adopt the same training and testing sequences with \cite{li2019net,li2020dmlo}. Compared with \cite{li2020dmlo}, our method utilizes soft correspondence rather than exact matching pairs so as to relize end-to-end pose estimation, and we achieved similar results on $t_{rel}$ and better results on $r_{rel}$. Compared with \cite{li2019net}, our method does not need an extra mask network and can filter the outliers with the proposed hierarchical embedding mask. Moreover, our method does not need to project 3D point clouds to 2D \cite{li2019net,li2020dmlo} and obtains the LiDAR odometry directly from the raw 3D point clouds. We achieved better results than LO-Net~\cite{li2019net}, even than LOAM~\cite{zhang2017low} on most sequences. We believe the pose refinement target makes our internal trainable mask effective for various outliers other than only dynamic regions~\cite{li2019net} as shown in Fig.~\ref{fig:visual}. Moreover, the trainable iterative refinement makes the estimated pose refined multiple times in one network inference.

\vspace{0.1cm}

\setlength{\tabcolsep}{0.9mm}
\begin{table*}[t]
	\footnotesize
	\begin{center}
		\resizebox{1.0\textwidth}{!}
		{
			\begin{tabular}{l|l||cc|cc|cc|cc|cc|cc|cc|cc|cc|cc|cc||cc}
				\toprule
				& &  \multicolumn{2}{c|}{00$^*$}  &\multicolumn{2}{c|}{01$^*$}      & \multicolumn{2}{c|}{02$^*$} & \multicolumn{2}{c|}{03$^*$} &  \multicolumn{2}{c|}{04$^*$} & \multicolumn{2}{c|}{05$^*$} & \multicolumn{2}{c|}{06$^*$} & \multicolumn{2}{c|}{07} & \multicolumn{2}{c|}{08} & \multicolumn{2}{c|}{09} &\multicolumn{2}{c||}{10} &\multicolumn{2}{c}{Mean on 07-10} \\ 
				\cline{3-26}\noalign{\smallskip}
				
				& \multirow{-2}{*}{\begin{tabular}[c]{@{}c@{}}Method \end{tabular}}
				&  $t_{rel}$  & $r_{rel}$   & $t_{rel}$                       & $r_{rel}$               & $t_{rel}$                          & $r_{rel}$   & $t_{rel}$ & $r_{rel}$   & $t_{rel}$                          & $r_{rel}$   & $t_{rel}$ & $r_{rel}$    & $t_{rel}$                          & $r_{rel}$   & $t_{rel}$ & $r_{rel}$ & $t_{rel}$                          & $r_{rel}$   & $t_{rel}$ & $r_{rel}$    & $t_{rel}$ & $r_{rel}$  & $t_{rel}$ & $r_{rel}$       \\
				\hline\hline
				\noalign{\smallskip}
				(a)   
				&Ours w/o mask    
				&0.99 & 0.58
				&0.61 & 0.25	
				& 1.07 & 0.55 
				& 0.84 &  0.66
				& 0.42 &  0.58
				& 0.68 &  0.41
				& 0.33 &  0.25
				& 0.76 & 0.58
				& 1.54 &  0.66
				& 1.30 &  0.61
				& 2.36 & 0.86 
				& 1.490 & 0.678  
				\\
				& Ours w/o mask optimazation   
				& 0.91& 0.54
				&\bf0.59 & 0.26
				& 1.10 & 0.49 
				& 0.81 &  \bf0.43
				& 0.41 &  1.05
				&0.57  &  0.39
				& 0.34 &  \bf0.22
				& 0.67 & 0.54
				& 1.41 &  0.59
				& \bf0.78 &  0.39
				& \bf1.32 & 0.66 
				& \bf1.045  &  0.545 
				\\
				
				&Ours (full, with mask and mask optimazation)     
				&\bf0.78 & \bf0.42
				&0.67 & \bf0.23
				&\bf0.86 & \bf0.41
				&\bf0.76 & 0.44
				&\bf0.37 & \bf0.40
				&\bf0.45 & \bf0.27
				&\bf0.27 & \bf0.22
				&\bf0.60 & \bf0.44
				&\bf1.26 &  \bf0.55
				&0.79 & \bf0.35
				&1.69 & \bf0.62
				&1.085  &\bf 0.490   
				\\
				\cline{1-26}\noalign{\smallskip}

				(b) &Ours (w/o cost volume)    
				& 42.52 & 19.44
				& 22.73 & 4.44
				& 37.53  &  12.55
				& 22.21 &  13.18
				& 15.38 &  3.59
				& 31.95 &  12.88
				& 12.82 & 3.83
				& 30.78 & 18.95
				& 67.08 &  25.72
				& 38.63 &  16.38
				& 40.55  &  19.80
				& 44.260 & 20.213
				\\
				
				&Ours (with the cost volume in \cite{liu2019flownet3d})    
				&1.11 & 0.51
				&0.61 & \bf0.22
				&0.97  &  0.47
				& \bf0.68 &  0.61
				& 0.50 &  0.67
				& 0.51 &  0.30
				& 0.38 & \bf 0.17
				& 1.18 & 0.73
				& 1.40 &  0.59
				& 1.15 &  0.46
				&\bf1.50  &  0.88
				& 1.308 & 0.665
				\\
				&Ours (with the cost volume in \cite{wu2019pointpwc})      
				&\bf0.69 &\bf 0.35
				&\bf0.59 & 0.23
				& 0.87 &  0.42
				& 0.79 &  0.55
				&\bf 0.34 & \bf 0.34
				& 0.49 &  0.28
				& 0.28 &  \bf0.17
				& 0.62 &  0.47
				& 1.47 & 0.62
				& 0.97 &  0.48
				&1.54  & \bf0.59 
				& 1.150 &  0.540 
				\\
				&Ours (full, with the cost volume in \cite{wang2020hierarchical})     
				&0.78 & 0.42
				&0.67 & 0.23
				&\bf0.86 & \bf0.41
				&0.76 & \bf0.44
				&0.37 & 0.40
				&\bf0.45 & \bf0.27
				&\bf0.27 & 0.22
				&\bf0.60 & \bf0.44
				&\bf1.26 & \bf 0.55
				&\bf0.79 & \bf0.35
				&1.69 & 0.62
				&\bf1.085  &\bf 0.490  
				\\
				\cline{1-26}\noalign{\smallskip}

				(c) 
				&Ours w/o Pose Warp-Refinement    
				&15.71 & 8.26
				&13.92 & 5.62	
				& 13.38 &  7.07
				& 15.74 &  16.62
				& 5.00 &  6.89
				& 14.19 &  7.87
				& 6.38 &  5.33
				& 12.52 & 11.73
				& 15.64 &  9.11
				& 10.17 &  7.60
				& 16.09 &  11.29
				& 13.605 &  9.933 
				\\
				&Ours w/o Pose Warp    
				&3.07 & 1.40 
				&4.04 & 1.56 
				&3.27 & 1.27
				&2.56 & 2.01
				&0.91 & 1.13 
				&2.28 & 1.15 
				&2.08 & 1.18 
				&4.29 & 2.75
				&5.00 & 2.14 
				&4.48 & 2.19
				&5.39 & 3.36
				&4.790 &  2.610
				\\
				
				&Ours (full, with Pose Warp-Refinement)     
				&\bf0.78 & \bf0.42
				&\bf0.67 & \bf0.23
				&\bf0.86 & \bf0.41
				&\bf0.76 & \bf0.44
				&\bf0.37 & \bf0.40
				&\bf0.45 & \bf0.27
				&\bf0.27 & \bf0.22
				&\bf0.60 & \bf0.44
				&\bf1.26 & \bf 0.55
				&\bf0.79 & \bf0.35
				&\bf1.69 & \bf0.62
				&\bf1.085  &\bf 0.490     
				\\
				
				\cline{1-26}\noalign{\smallskip}
				(d) 
				&Ours (first embedding on the last level)
				& 0.95 &0.46 
				&\bf0.67 & 0.29
				& 1.03 & 0.44 
				& 0.96 &  0.77
				& 0.48 & \bf0.40
				& 0.74 &  0.40
				& 0.41 &  0.26
				& 0.94 & 0.49
				& 1.41 &  0.56
				& 1.01 &  0.47
				&\bf1.67  &  0.82
				& 1.258 &   0.585
				\\
				&Ours (full, first embedding on the penultimate level)     
				&\bf0.78 & \bf0.42
				&\bf0.67 & \bf0.23
				&\bf0.86 & \bf0.41
				&\bf0.76 & \bf0.44
				&\bf0.37 & \bf0.40
				&\bf0.45 & \bf0.27
				&\bf0.27 & \bf0.22
				&\bf0.60 & \bf0.44
				&\bf1.26 &  \bf0.55
				&\bf0.79 & \bf0.35
				&1.69 &\bf 0.62
				&\bf1.085  &\bf 0.490  
				\\
				\bottomrule
			\end{tabular}
		}
	\end{center}
	\vspace{-8pt}
	\caption{The ablation study results of LiDAR odometry on KITTI odometry dataset \cite{geiger2013vision}.}
	\vspace{-9pt}			
	\label{table:ablation}
\end{table*}

\begin{figure}[t]
	\begin{center}
		\resizebox{0.95\columnwidth}{!}
		{
			\includegraphics[scale=1.00]{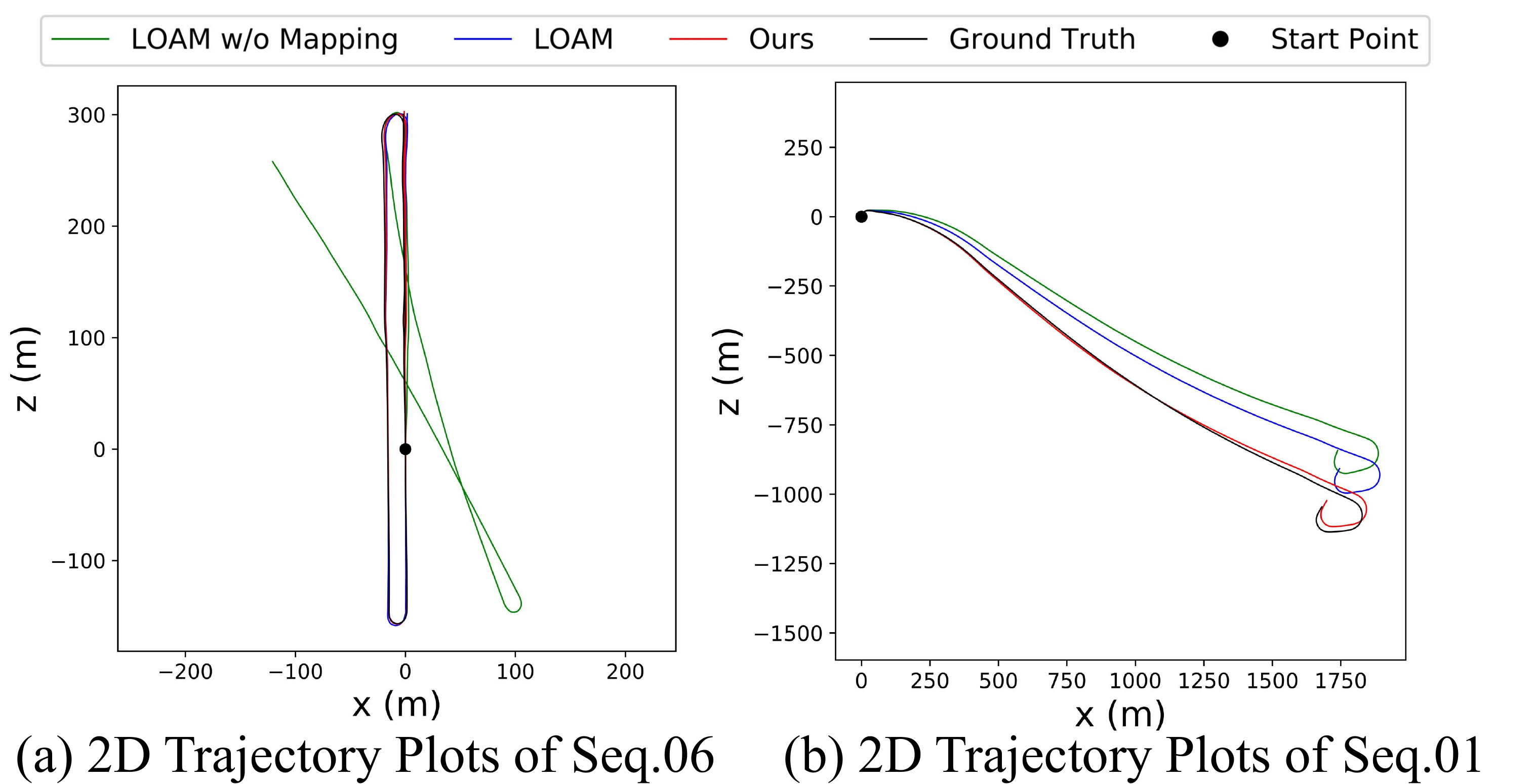}}
	\end{center}
	\vspace{-8pt}
	\caption{Trajectory results of LOAM and ours on KITTI training sequences with ground truth. Ours is much better than the LOAM without mapping. And ours also has better performance on the two sequences than full LOAM, though ours is for odometry and LOAM has the mapping optimization.}
	\label{fig:odometry_path2d}
	\vspace{-8pt}
\end{figure}

\begin{figure}[t]
	\begin{center}
		\resizebox{0.95\columnwidth}{!}
		{
			\includegraphics[scale=1.00]{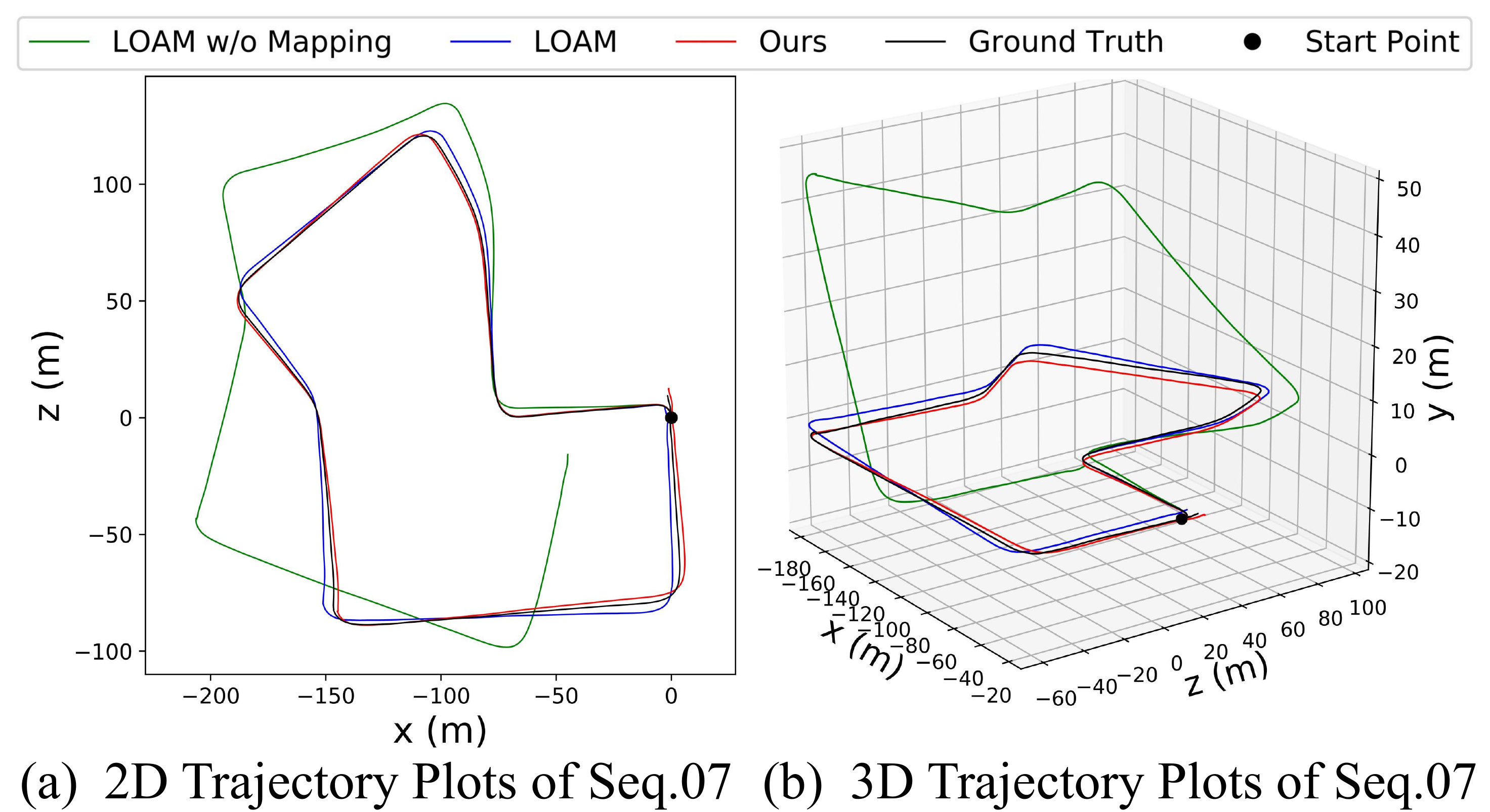}}
	\end{center}
	\vspace{-8pt}
	\caption{3D and 2D trajectory results on KITTI validation sequence 7 with ground truth. Our method obtained the most accurate trajectory.}
	\label{fig:odometry_path2d3d}
	\vspace{-9pt}
\end{figure}

\begin{figure}[t]
	\begin{center}
		\resizebox{0.95\columnwidth}{!}
		{
			\includegraphics[scale=1.00]{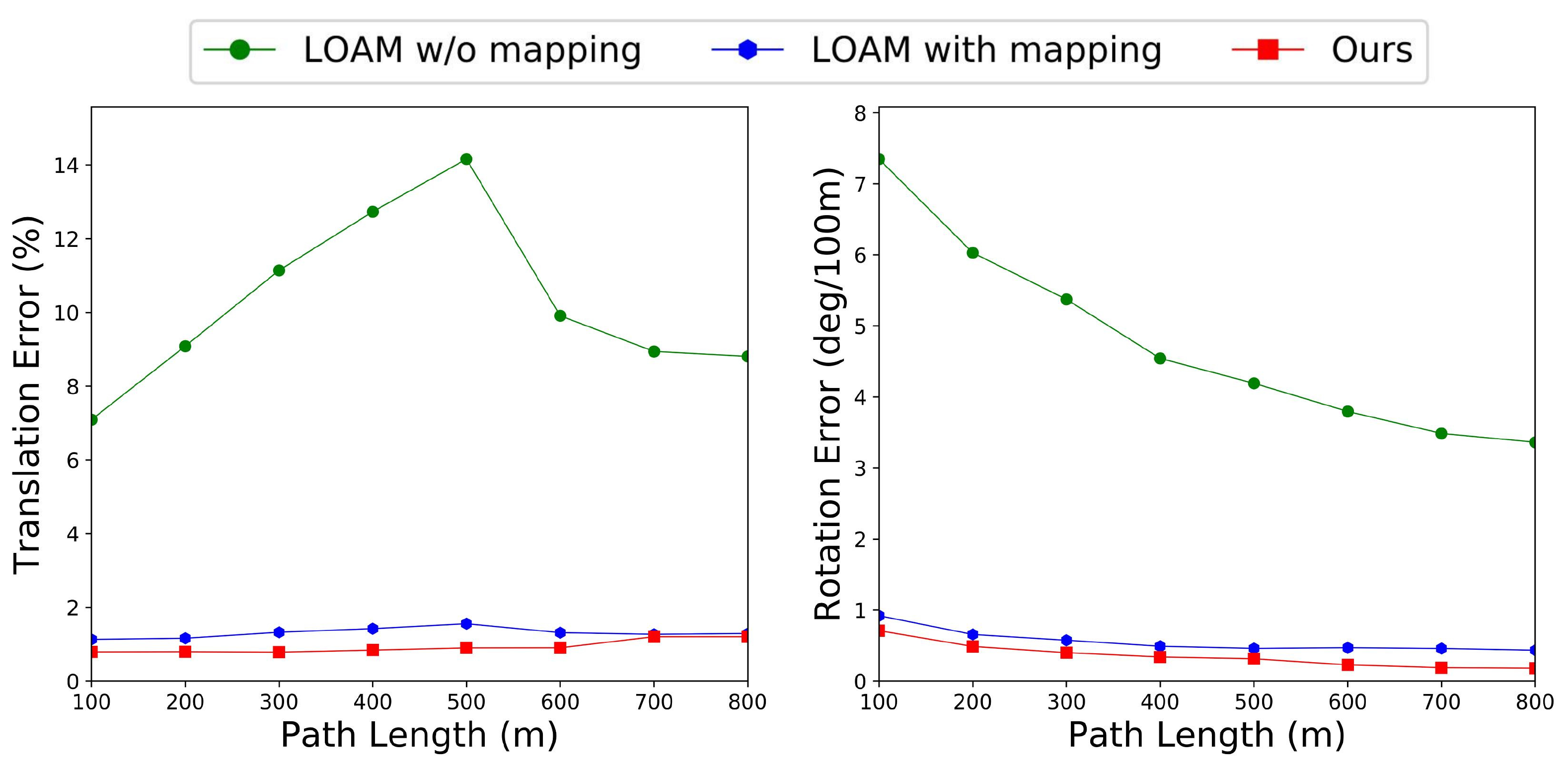}}
	\end{center}
	\vspace{-8pt}
	\caption{Average translational and rotational error on KITTI sequences 00-10 on all possible subsequences in the length of $100,200,...,800m$. Our method has the best performance.}
	\label{fig:odometry_error_line}
	\vspace{-8pt}
\end{figure}

\vspace{-0.05cm}
\noindent{}{\bf Using other sequences as training/testing sets: }

\noindent{}{\bf 00-06, 09-10/07-08:} Quantitative results are listed in Table~\ref{table:lidar_7_8} to compare with a recent method \cite{zheng2020lodonet}.

\noindent{}{\bf 00-03, 05-09/04, 10:} As shown in Table~\ref{table:lidar4_10}, we adopted the same training/testing sets of the KITTI odometry dataset \cite{geiger2013vision} and compared with \cite{wang2019deeppco}. 

\noindent{}{\bf 00-08/09-10:} \cite{cho2020} proposes an unsupervised method on LiDAR odometry. Table~\ref{table:lidar09_10} shows the quantitative comparison of this method with ours.

The results in Table~\ref{table:lidar_7_8}, Table~\ref{table:lidar4_10}, and Table~\ref{table:lidar09_10}  show that our method outperforms them. \cite{wang2019deeppco} is based on 2D convolutional network, which loses the raw 3D information. \cite{zheng2020lodonet} finds matched keypoint pairs in 2D depth images to regress pose. \cite{cho2020} is a method of unsupervised training. The results demonstrate the superiority of our full 3D learning based LiDAR odometry. 

The qualitative results are shown in Figs.~\ref{fig:odometry_path2d},~\ref{fig:odometry_path2d3d}, and~\ref{fig:odometry_error_line}. We compared our method with LOAM \cite{zhang2017low} and LOAM without mapping since \cite{li2019net,zheng2020lodonet,wang2019deeppco,li2020dmlo} do not release their codes. Full LOAM has a superb performance but degrades significantly without mapping. Ours is only for odometry and is better than the LOAM without mapping. At the same time, ours is even better than LOAM \cite{zhang2017low} on average evaluation.

\vspace{-0.15cm}
\subsection{Ablation Study}
\vspace{-4pt}

In order to analyze the effectiveness of each module, we remove or change components of our model to do the ablation studies on KITTI odometry dataset. The training/testing details are the same as described in Sec.~\ref{sec:implementation}.

\vspace{5pt}
\noindent{}{\bf Benefits of Embedding Mask:}
We first remove the optimization of the mask, which means that the embedding masks are independently estimated at each level. Then we entirely remove the mask and apply the average pooling to embedding features to calculate the pose. The results in Table~\ref{table:ablation}(a) show that the proposed embedding mask and its hierarchical optimization both contribute to better results. 

\begin{figure}[t]

	\centering
	\resizebox{0.95\columnwidth}{!}
	{
		\includegraphics[scale=1.00]{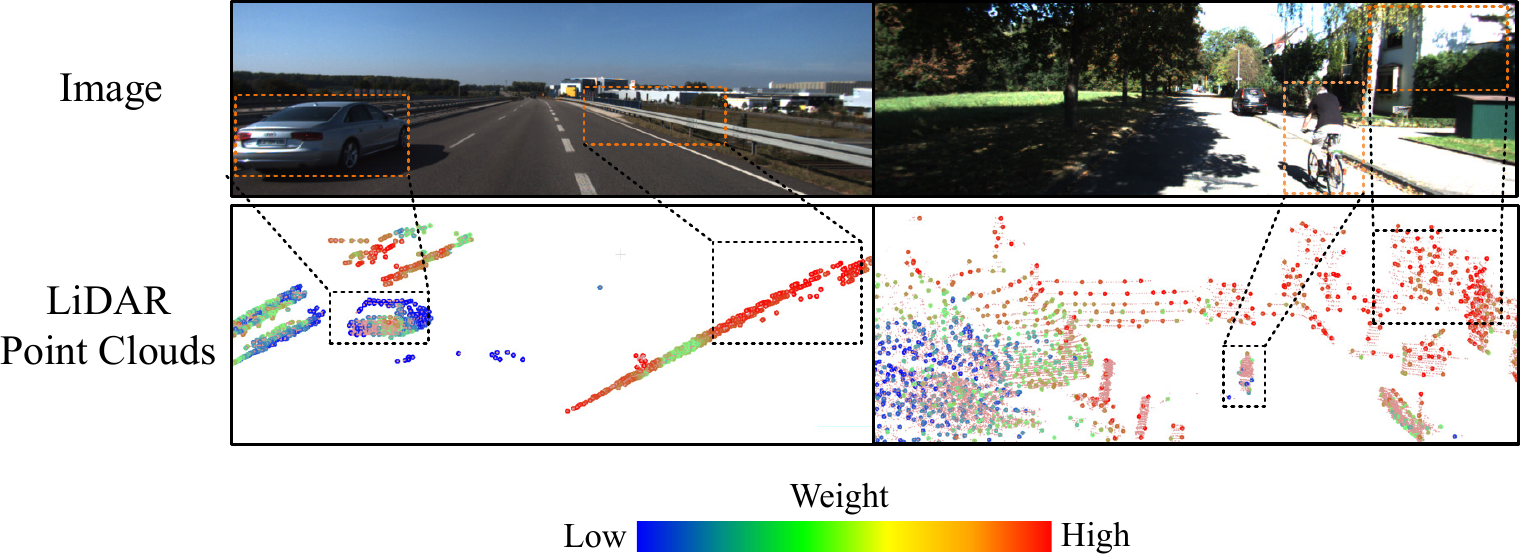}}
	\vspace{-0.1mm}
	\caption{The Visualization of embedding mask. In the LiDAR point clouds, the small red points are the whole point cloud of the first frame, and the big points with distinctive colors are the sampled 2048 points with contributions to the pose transformation. In the example on the left, the buildings and steel bars around the highway have high weights, while the bushes and the moving car have low weights. In the example on the right, the buildings, the tree trunks, and the static car have high weights, while the weeds and the cyclist have low weights.}
	\label{fig:visual}
    	\vspace{-0.4cm}
\end{figure}

The mismatched objects are different in various scenes. In some scenes, the car is dynamic, while the car is static in another scene (Fig.~\ref{fig:visual}). Through the learnable embedding mask, the network learns to filter outlier points from the overall motion pattern. The network achieves the effect of RANSAC~\cite{fischler1981random} by only once network inference.

\vspace{3pt}
\noindent{}{\bf Different Cost Volume:} 
As there are different point feature embedding methods in point clouds, we compare three recent methods in our odometry task, including the flow embedding layer in FlowNet3D \cite{liu2019flownet3d}, point cost volume~\cite{wu2019pointpwc} and the attentive cost volume~\cite{wang2020hierarchical}. The results in Table~\ref{table:ablation}(b) show that the model with double attentive cost volume has the best results in the three. Therefore, robust point association also contributes to the LiDAR odometry task.

\vspace{5pt}
\noindent{}{\bf Effect of Pose Warp-Refinement:}
We first remove the pose warping and reserve the hierarchical refinement of the embedding feature and the mask. The results show a decline as in Table~\ref{table:ablation}(c). Then we remove the full pose warp-refinement module, which means that the embedding features, mask, and pose are only calculated once. As a result, the performance degrades a lot, which demonstrates the importance of the coarse-to-fine refinement.

\vspace{5pt}
\noindent{}{\bf Suitable Level for the First Embedding:}
As fewer points have less structural information, it is needed to decide which level is used to firstly correlate the two point clouds and generate the embedding features. We test the different levels of the first feature embedding. The results in Table~\ref{table:ablation}(d) demonstrate that the most suitable level for the first embedding is the penultimate level in the point feature pyramid, which shows that the 3D structure is also important for the coarse pose regression.

	\vspace{-0.2cm}

\subsection{Visualization of Embedding Mask}
\vspace{-4pt}
The proposed network uses embedding features of 2048 points to calculate the pose transformation in the final pose output layer. We visualize the embedding mask in the 2048 points to show the contribution of each point to the final pose transformation. As illustrated in Fig.~\ref{fig:visual}, the points that are sampled from the static and structured rigid objects such as buildings, steel bars, and the static car have higher weights. On the contrary, the points from the dynamic and irregular objects such as the moving car, the cyclist, bushes, and weeds have lower weights. In LO-Net~\cite{li2019net}, only the dynamic regions are masked, while our method also gives small weight to unreliable weeds and bushes. Therefore, the embedding mask can effectively reduce the influence of dynamic objects and other outliers on the estimation of pose transformation from adjacent frames. \vspace{-0.2cm}

\section{Conclusion}
\vspace{-4pt}
Different from the 2D projection-based methods \cite{nicolai2016deep,velas2018cnn,wang2019deeppco,li2019net}, a full 3D learning method based on PWC structure for LiDAR odometry is proposed in this paper. The pose warp-refinement module is proposed to realize the hierarchical refinement of 3D LiDAR odometry in a coarse-to-fine approach. The hierarchical embedding mask optimization is proposed to deal with various outlier points. And our method achieved a new state-of-the-art end-to-end LiDAR odometry method with no need for an extra mask network \cite{li2019net}. Since our mask can filter a variety of outlier points that are not suitable for calculating overall motion, it is a direction worth exploring to combine our new mask with mapping optimization in the future.

\vspace{4pt}
{\bf \small  Acknowledgement.} {\small This work was supported in part by the Natural Science Foundation of China under Grant U1613218, U1913204, 62073222, in part by "Shu Guang" project supported by Shanghai Municipal Education Commission and Shanghai Education Development Foundation under Grant 19SG08, in part by grants from NVIDIA Corporation.}

{\small
\bibliographystyle{ieee_fullname}
\bibliography{egbib}
}

\end{document}